\documentclass[journal]{IEEEtran}
\usepackage{cite}
\usepackage{amsmath,amssymb,amsfonts}
\usepackage{algorithmic}
\usepackage{graphicx}
\usepackage{textcomp}
\usepackage{xcolor}
\usepackage{algorithm}

\usepackage{booktabs}       

\ifCLASSINFOpdf
\else
\fi
\begin{document}
%
\title{ES-dRNN: A Hybrid Exponential Smoothing and Dilated Recurrent Neural Network Model for Short-Term Load Forecasting}
%
%
%

\author{Slawek~Smyl,
	Grzegorz~Dudek,
	and~Paweł~Pełka

\thanks{S. Smyl works at Facebook, 1 Hacker Way, Menlo Park, CA 94025, USA,
e-mail: slawek.smyl@gmail.com}
\thanks{G. Dudek and P. Pełka are with the Department of Electrical Engineering, Czestochowa University of Technology, 42-200 Czestochowa, Al. Armii Krajowej 17, Poland, e-mail: grzegorz.dudek@pcz.pl, pawel.pelka@pcz.pl.}
\thanks{}}
	
%
%

\markboth{}%
{Shell \MakeLowercase{\textit{et al.}}: Bare Demo of IEEEtran.cls for IEEE Journals}
%



\maketitle

\begin{abstract}
	Short-term load forecasting (STLF) is challenging due to complex time series (TS) which express three seasonal patterns and a nonlinear trend. This paper proposes a novel hybrid hierarchical deep learning model that deals with multiple seasonality and produces both point forecasts and predictive intervals (PIs). It combines exponential smoothing (ES) and a recurrent neural network (RNN). ES extracts dynamically the main components of each individual TS and enables on-the-fly deseasonalization, which is particularly useful when operating on a relatively small data set. A multi-layer RNN is equipped with a new type of dilated recurrent cell designed to efficiently model both short and long-term dependencies in TS. To improve the internal TS representation and thus the model's performance, RNN learns simultaneously both the ES parameters and the main mapping function transforming inputs into forecasts.
	We compare our approach against several baseline methods, including classical statistical methods and machine learning (ML) approaches, on STLF problems for 35 European countries. The empirical study clearly shows that the proposed model has high expressive power to solve nonlinear stochastic forecasting problems with TS including multiple seasonality and significant random fluctuations. In fact, it outperforms both statistical and state-of-the-art ML models in terms of accuracy.

\end{abstract}


%
\IEEEpeerreviewmaketitle

\section{Introduction}
\IEEEPARstart{E}{}lectricity demand forecasting for different horizons and granularity is an integral part of power system control, scheduling and planning. Thus, it is extremely important for energy suppliers, system operators, financial institutions, and other participants in electric energy generation, transmission, distribution, and markets. At a short-term level, i.e. with a horizon from one hour to seven days ahead and hourly granularity or less, electricity demand forecasting is the basis of power system operation including unit commitment, generation dispatch, hydro scheduling, hydrothermal
coordination, spinning reserve allocation, interchange and
low flow evaluation, security assessment, and network diagnosis \cite{Dud19}. Modern power systems pose new challenges for the forecasting models due to issues connected with volatile distributed energy resources, integration of intermittent renewable energy resources and deployment of demand-side management.  
As electricity demand is the primary driver of electricity prices, STLF plays a key role in competitive energy markets. The accuracy of forecasts translates directly into the financial performance of energy market participants.
A related study revealed that 
a 1\% reduction in forecasting error for a 10 GW utility can save up to \$1.6 million annually \cite{Hob95}.

\subsection{Related Work}

The importance of STLF for the safe, reliable and efficient operation of power systems as well as the complexity of the problem translates into great interest from researchers in this field. Nonlinear trend, multiple seasonality, variable variance and daily profile, and random fluctuations, make STLF challenging and place high demands on forecasting models.
STLF approaches can be divided into three categories: statistical or econometric models, ML models, and hybrid ones. The first category includes auto-regressive integrated moving average (ARIMA) \cite{Aro18}, exponential smoothing (ES) \cite{Tay12}, linear regression \cite{Cha14}, and Kalman filtering \cite{Tak16}. 

The main problem with statistical STLF methods is their linear nature, which limits the implementation of non-linear system dynamics. To extend the model's capabilities to approximate nonlinear relationships, local modeling is used. For example in \cite{Sha20} the linear state-space model is learnt progressively from the data using a Kalman filter. The method works by assuming temporally local linearity, which can be seen as an approximation of an underlying nonlinearity, generalizing the standard linear-Gaussian model with static parameters. In \cite{Dud16} the target nonlinear function was modeled locally in the neighborhood around the query pattern using linear regression. Due to initial data normalization, which simplified relationships between input and output data, linear models, such as partial least-squares regression, were able to compete with more sophisticated ML models.

Another problem with statistical methods is their limited ability to model complicated seasonal patterns. Standard ARIMA and Holt-Winters models can be extended to multiple seasonality \cite{Tay10} but they assume that the cycle shapes are all the same. In practice, the seasonal patterns can greatly differ from each other (in STLF, daily cycles for workdays are usually significantly different from those for weekends). Taking into account the changing seasonal pattern requires a significant extension of the model. For example, in \cite{Liv11}, an ES state space model was combined with Fourier terms, a Box-Cox transformation and ARMA error correction.
Extending the regression model with Fourier terms (harmonic regression) is a popular method of introducing seasonal components into statistical models \cite{Tay18}. Another approach to deal with seasonality is TS decomposition. Products of decomposition are less complex than the original TS and can be modeled using simpler models \cite{Dud16}, \cite{Fan12}, \cite{Hov15}. 

Other drawbacks statistical methods suffer from are limited adaptability, a shortage of expressive power, problems with capturing long-term dependencies and introducing exogenous variables into the model. ML methods offer many more possibilities than statistical ones. 
They provide forecasting models with the ability to learn historical patterns and anomalies and successively improve prediction accuracy. The most researched ML models in the field of forecasting are neural networks (NNs) \cite{Ben20}. They can flexibly model complex nonlinear relationships between variables and reflect process variability in uncertain dynamic environments due to their universal approximation property. 
At the same time, NNs have their limitations such as disruptive and unstable training, need for careful feature engineering, local optimality, weak interpretability, difficulty in matching the network architecture to the problem solved, tendency to overfitting, weak extrapolation ability and many parameters to estimate.
These issues as well as problems with modeling complex seasonal patterns are addressed in STLF literature in various ways. For example in \cite{Dud16a}, TS with multiple seasonality were represented by patterns of the daily profiles, which simplified greatly the forecasting problem. As a result, it was possible to use simpler, resistant to overfitting neural models with a small number of parameters. Among the NN architectures compared in \cite{Dud16a} are multilayer perceptron (MLP), radial basis function (RBF) NN, generalized regression neural network (GRNN), fuzzy counterpropagation NN, and self-organizing maps. Within the group, GRNN and MLP turn out to be the most accurate. A Bayesian approach was used in \cite{Hip10} to control MLP complexity and to select input variables. The Bayesian framework offered ways to avoid overfitting by regularisation, to decide on the number of neurons by comparing the model evidences, and to deal with the inputs by soft-pruning. Many NN solutions for STLF combine the neural model, optimization method for hyperparameter selection and learning, and TS decomposition or a feature engineering method. An example can be found in \cite{Bas09} where TS is decomposed using wavelet transform to extract relevant information from the load curve, and MLP weights are adjusted using a particle swarm optimization algorithm. The most popular and universal NN, MLP, was recently replaced by randomized NN in STLF \cite{Dud21a}. When a pattern-based representation is used, randomized NN can produce more accurate forecasts than MLP while having many advantages over MLP. These include extremely fast and easy training, simple architecture, small number of hyperparameters and parameters to estimate, and ease of implementation.

NN architectures proposed for STLF in recent years are dominated by deep learning (DL) and recurrent NNs (RNN). The success of DL can be largely attributed to increased model complexity and the ability to cross-learn on massive datasets. This strengthens expressive power and the ability to extract patterns across multiple examples. DL architectures are
composed of combinations of basic structures, such as MLPs, convolutional NNs (CNNs) and RNNs. New ideas in the field of DL have been successfully applied to STLF. Some examples are: \cite{Che19}, where deep residual NNs were proposed and applied to probabilistic load forecasting using Monte Carlo dropout; \cite{Hos19}, where a multivariate fuzzy TS was converted into multi-channel images and processed by CNN to produce load forecasts; \cite{Kon20}, where an improved deep belief network for STLF considering demand-side management was proposed; and \cite{Kon19}, where a STLF problem for individual residential households was addressed using long-short term memory RNN (LSTM). Among NN forecasting models, modern RNNs such as LSTM and gated recurrent unit (GRU) are distinguished by their ability to model both short and long-term dependencies in TS. Therefore they are readily used for STLF \cite{Wan19}. 

To improve further forecasting model performance, ensemble and  hybrid solutions have been developed. Ensembling is a reliable approach to increase the forecast accuracy and robustness of both statistical and ML models. It combines, in some way, multiple models to produce a common response, controlling a bias-variance-covariance trade-off \cite{Bro05}. Ensemble strategies for STLF take many forms. For example in \cite{ElH20}, an ensemble of NNs is proposed, which are trained on the products of wavelet decomposition; in \cite{Yan22} an empirical mode decomposition is applied to decompose the original interval-valued STLF data, and an ensemble of LSTMs is utilized to synchronously forecast and combine the components; and in \cite{Hu21} a data-driven multi-objective evolutionary ensemble learning is proposed with Random Vector Functional Link NNs as base learners. Hybrid approaches combine two or more methods in a common model, taking advantage of their strengths and avoiding their drawbacks. For example, statistical methods can help in data preprocessing and reduce overfitting of ML models.
Examples of model hybridization for STLF can be found in \cite{Wan21} where a temporal CNN is utilized to extract hidden information and long-term temporal relationships in the input data and a boosted tree model (LightGBM) is used to predict future loads based on the extracted features, and in \cite{Lai21} where both LSTM and wavelet decomposition extract TS features, on which an ensemble of RBF NNs is trained. The produced forecasts are aggregated by a localized generalization error model, which optimizes the ensemble member weights.

\subsection{Motivation and Contribution} 

The motivation behind this work is threefold. First, STLF is extremely important for power system operation and energy market functioning. Forecast accuracy directly translates into the safe, reliable and efficient operation of power systems as well as improved financial performance of energy market participants. Second, STLF is a challenging problem due to complex three component seasonality, nonlinear trend, significant stochastic component and changing seasonal patterns. It requires a flexible forecasting model capable of capturing long-term and short-term dependencies in TS. Third, new advances in ML and DL, especially in sequential data processing, encourage their application in complex forecasting problems such as STLF. Modern RNNs can deal with multiple seasonality and long-term dependencies in TS. Hybrid solutions utilizing both statistical and DL methods improve representation learning and the exploration of hidden patterns. In this study, we extend our recent works \cite{Smy20} and \cite{Dud21}, where we used a combined ES and LSTM model for forecasting TS with single and double seasonality. It is worth noting that the hybrid model proposed in \cite{Smy20} 
won the renowned M4 forecasting competition in 2018, outperforming a wide variety of state-of-the-art models.
The winning model produced both the most accurate forecasts and the most precise PIs for 100,000 real-life TS. 
It was close to 10\% more accurate than the benchmark ensemble model, which is a huge improvement \cite{Mar18}. 
This means, the model has been reliably and rigorously verified on a wide range of forecasting problems. In this study, based on the main concept of the winning solution, we develop a new model specifically for an STLF problem with three seasonal patterns. 



Our research contributions can be summarized as follows:

\begin{enumerate}
\item We propose a new dilated recurrent cell, dRNNCell, as a building block of dilated RNNs designed especially for STLF to deal with both short and long-term dependencies in TS. 
\item We develop a new hybrid forecasting model for STLF combining ES and RNN. The model produces point and probabilistic forecasts in the form of PIs. It does not require initial TS decomposition and, due to its internal mechanisms such as adaptive TS preprocessing, cross-learning and multiple dilation, can deal with complex TS expressing nonlinear trend, varying variance and multiple seasonality.    
\item We propose a new mechanism for dynamically adjusting the smoothing coefficients used by ES. These coefficients are learned by RNN simultaneously with the main mapping function transforming inputs into forecasts to ensure optimal internal representation of TS and finally maximize the accuracy of the model.
\item We introduce a new three-component loss function based on pinball loss to optimize both the point forecasts and PIs,   
\item We empirically demonstrate on real-world data for 35 European countries that the proposed hybrid model outperforms in STLF well-established statistical and state-of-the-art ML approaches.

\end{enumerate}

The rest of the work is organized as follows. Section II describes the STLF data and defines the STLF problem. Section III presents the proposed forecasting model: its architecture, components and features. 
The experimental framework used to evaluate the proposed model is described in Section IV. Finally, Section V concludes the work.

\section{STLF Problem and Data}

\begin{figure}
	\centering
	\includegraphics[width=0.24\textwidth]{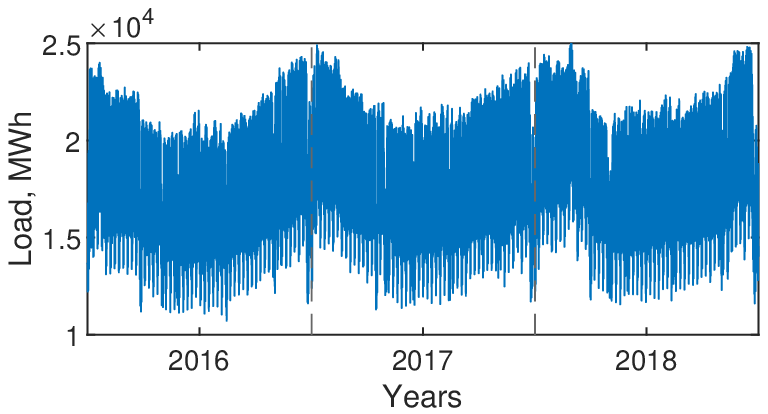}
	\includegraphics[width=0.24\textwidth]{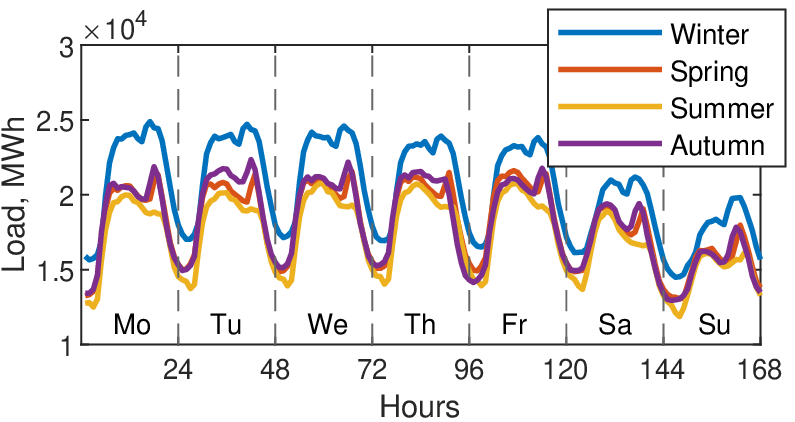}
	\caption{Hourly electricity demand TS for Poland.} 
	\label{figPL}
\end{figure}

In this study, we consider a univariate STLF problem where the task is to forecast future values of the hourly electricity demand TS for the next day, $\{{z}_\tau\}_{\tau=M+1}^{M+24}$, given a sequence of past observations, $\{z_\tau\}_{\tau=1}^{M}$. The problem is challenging because the electricity demand TS exhibits a trend, three types of variability: the daily, weekly and yearly ones, and random fluctuations -- see Fig. \ref{figPL}, where hourly electricity demand for the Polish power system is shown.
The level of electricity demand and its long-term trend depend on a country's economic development and growth rate. 
One of the most important factors which can upset electricity demand in the short-term perspective are extreme weather conditions.

\begin{figure}
	\centering
	\includegraphics[width=0.48\textwidth]{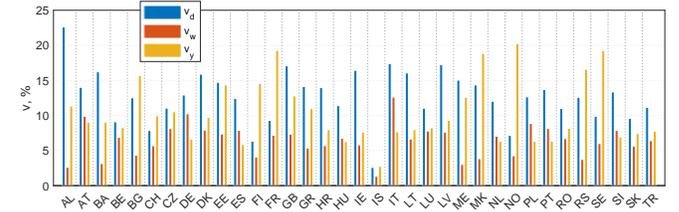}
	\caption{Coefficients of daily, weekly and yearly variations of electricity demand.} 
	\label{figVc}
\end{figure}

A very important issue from the point of view of power system control and planning is electricity demand variability observed in daily, weekly and yearly periods. Fig. \ref{figVc} shows variation coefficients for 35 European countries defined as $v=100s/\bar{z}$. For daily electricity demand variations, $v_d$, $\bar{z}$ and $s$ express the daily mean and standard deviation, for weekly variations, $v_w$, express the weekly mean and standard deviation of daily means, and for yearly variations, $v_y$, express the yearly mean and standard deviation of weekly means. Greater demand variability requires greater flexibility of generating units and the entire power system. As can be seen from Fig. \ref{figVc}, the lowest demand variations are for Iceland. The strongest daily variations ($v_d>17\%$) are for Albania, Italy, Latvia and Great Britain, while the strongest yearly variations ($v_y>18\%$) are for Norway, France, Sweden and 	Macedonia. The weekly variations are usually weaker than the daily and yearly ones, $v_w<13\%$. Countries with the strongest weekly variations are Italy, Germany, Austria and Poland. It is worth noting that the electricity demand variation changes over time, which is an additional challenge for forecasting models.  

\begin{figure}
	\centering
	\includegraphics[width=0.48\textwidth]{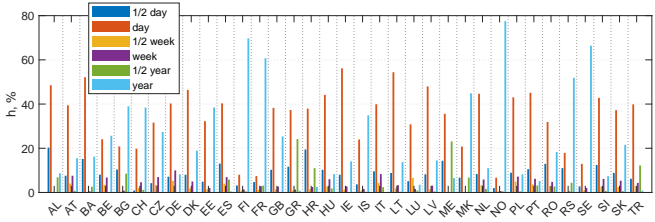}
	\caption{Contribution of the most important harmonics in the TS variance.} 
	\label{figHm}
\end{figure}

The electricity demand seasonalities are related to local climate, weather variability and the types of consumers. Intensities of the seasonal fluctuations can be identified using harmonic analysis. Based on Parseval's theorem, the variance of a TS can be expressed by the sum of squares of its harmonic amplitudes. The contribution of the $i$-th harmonic to the variance can be expressed by the ratio $h_i=100 A_i^2/(2Var(z_t))$, where $A_i$ is the $i$-th harmonic amplitude and $Var(z_t)$ is the TS variance. Fig. \ref{figHm} shows ratio $h_i$ for the most important harmonics for 35 European countries. Note that for some countries the yearly seasonality strongly dominates compared to others ($h>60\%$). These countries include Finland, France, Norway and Sweden.
Another extremely important seasonality is the daily one. The countries with the strongest daily seasonality ($h>50$) are Bosnia and Herzegovina, Ireland and Lithuania. In contrast, the weekly seasonality is less distinct with $h$ below $10\%$ for all countries. The highest weekly fluctuations ($h>7\%$) are shown by Germany, Italy, Austria and Poland. Some countries demonstrate stronger half-yearly seasonality than  yearly ones. They include southern European countries such as Spain, Greece, Croatia, Italy, Montenegro and Turkey. The half-yearly seasonality is related to the tourism industry, which increases energy demand in the summer season.

\begin{figure}
	\centering
	\includegraphics[width=0.48\textwidth]{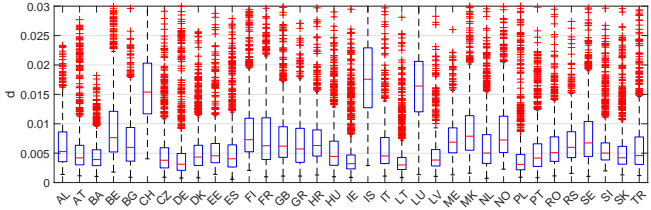}
	\caption{Boxplots of distances between daily patterns of electricity demand.} 
	\label{figDs}
\end{figure}

As can be seen from Fig. \ref{figPL}, the daily patterns for Tuesday through Friday from the same period of the year are similar, while those for Monday, Saturday and Sunday are distinct. The daily shapes are dependent on the period of the year and can vary over the years. High similarity in daily shapes makes forecasting easier. Fig. \ref{figDs} shows boxplots for distances between daily patterns representing the same days of the neighboring weeks, $d_t= \Vert \hat{\textbf{z}}_t - \hat{\textbf{z}}_{t+7} \Vert_2$. The daily pattern is defined as the centered and normalized daily vector: $\hat{\textbf{z}}_t = (\textbf{z}_t - \bar{z}_t)/\Vert(\textbf{z}_t - \bar{z}_t)\Vert_2$, where $\textbf{z}_t = [z_{t,1}, ..., z_{t,24}]$ is the vector of hourly demands for the $t$-th day and $\bar{z}_t$ is the mean demand for that day. Vectors $\hat{\textbf{z}}_t$ have mean zero, the same variance and unity length. The daily profiles expressed by $\hat{\textbf{z}}_t$ differ only in shape. From Fig. \ref{figDs} we can observe that the most similar profiles are for Lithuania, Poland, Germany and Ireland ($0.0030<d<0.0033$), while the most dissimilar ones are for Iceland, Luxembourg and Switzerland ($0.015<d<0.018$).          

\section{Forecasting Model}

\begin{figure}
	\centering
	\includegraphics[width=0.45\textwidth]{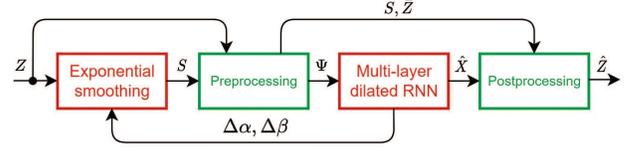}
	\caption{Block diagram of the proposed forecasting system.} 
	\label{figBd}
\end{figure}

A block diagram of the proposed forecasting model is shown in Fig. \ref{figBd}. The model is trained in a cross-learning mode \cite{Smy20}, i.e. simultaneously on $L$ hourly electricity load TS. Input $Z$ represents a set of $L$ TS: $\{\{z_\tau^l\}_{\tau=1}^{M_l}\}_{l=1}^L$, where $M_l$ is an $l$-th TS length. Output $\hat{Z}$ is a set of $L$ forecasts of the daily sequences $\{\{\hat{z}_\tau^l\}_{\tau=M_l+1}^{M_l+24}\}_{l=1}^L$. An exponential smoothing component expresses each TS from $Z$ by two smoothing equations, i.e. for level and seasonality. The seasonal components, $S$, are used by the prepossessing component to deseasonalize the TS. This component also normalizes and squashes the TS and prepares training sets $\Psi$ for RNN learning. It feeds the processing parameters, i.e.  seasonal components $S$ and average values of TS sequences $\bar{Z}$, to the postprocessing component. RNN produces forecasts for the deseasonalized, normalized and squashed daily sequences and their PIs for each TS ($\hat{X}$). These forecasts are postprocessed to obtain forecasts in real values, $\hat{Z}$. RNN also produces  corrections of the ES smoothing parameters, $\Delta\alpha$ and $\Delta\beta$, to tune them properly. In the ensemble version, the model is trained $E$ times and the forecasts are averaged. The diversity of ensemble learners, which decides about ensemble learning success \cite{Bro05}, is achieved by random initial parameters.    

Details of the model are described below.  

\subsection{Exponential Smoothing Component}

As shown in Section II, the hourly electricity load TS, $\{z_\tau\}_{\tau=1}^M$, exhibits complex phenomena with three seasonalities. To deal with the  challenging STLF problem, the TS is deseasonalized, normalized and squashed. Then, it is predicted by RNN. Deseasonalization is performed using a seasonal component produced by a simplified Holt-Winters multiplicative seasonal model in the form:

\begin{equation}
\begin{aligned}
l_{\tau}=\alpha \frac{z_\tau}{s_{\tau}} + (1-\alpha)l_{\tau-1} \\
s_{\tau+168}=\beta \frac{z_\tau}{l_{\tau}} + (1-\beta)s_\tau
\label{eqls}
\end{aligned}
\end{equation} 
where $l_\tau$ is a level component, $s_\tau$ is a weekly seasonal component, and $\alpha$, $\beta \in [0, 1]$ are smoothing coefficients.

The series are hourly, and exhibit daily, weekly, and yearly
seasonalities. The (1) includes only weekly seasonality. However, the daily seasonality (24 hourly values) is part of the
weekly seasonality (168 hourly values). Additionally, the series are processed in 24-hour steps, so the daily seasonality is
to some extend ”escaped” - the RNN always learns to forecast 
a whole day starting from midnight. The yearly seasonality
impact is dealt with partly by normalization and partly by using
date-related regressors, month and week of the year

A unique feature of the model is that the smoothing coefficients are learned by RNN. In addition to predicting the TS sequence and its PI, RNN also predicts corrections for smoothing coefficients, $\Delta\alpha_t$ and $\Delta\beta_t$. 
The smoothing coefficients are adapted in each recursive step $t$ using the following corrections:

\begin{equation}
\begin{aligned}
\alpha_{t+1} = \sigma(I\alpha + \Delta\alpha_t)\\
\beta_{t+1} = \sigma(I\beta + \Delta\beta_t)
\label{eqab}
\end{aligned}
\end{equation} 
where $I\alpha$, $I\beta$ are initial values of the smoothing coefficients (hyperparameters), and $\sigma$ is a sigmoid function, which maintains the coefficients within a range from 0 to 1. 

The smoothing coefficients have a dynamic character. This is because the corrections produced by RNN in each recursive step depend on  current and past TS characteristics (shape, level, seasonal pattern) and time variables that indicate in what phase of the weekly, monthly and yearly cycles the predicted daily sequence is (see RNN input pattern \eqref{eqxp}).  The dynamic Holt-Winters equations take the form 
\begin{equation}
\begin{aligned}
l_{t,\tau}=\alpha_t \frac{z_\tau}{s_{t,\tau}} + (1-\alpha_t)l_{t,\tau-1} \\
s_{t,\tau+168}=\beta_t \frac{z_\tau}{l_{t,\tau}} + (1-\beta_t)s_\tau
\label{eqls1}
\end{aligned}
\end{equation} 





\subsection{Preprocessing and Postprocessing Components}

To prepare input and output data for RNN we use two adjacent moving windows: input window $\Delta^{in}$ of size 168 hours and output window $\Delta^{out}$ of size 24 hours. The input window covers a weekly period to expose the RNN to the specific features of the series in this period directly. The output window covers the forecasted daily sequence.

\begin{figure}
\centering
	\includegraphics[width=0.48\textwidth]{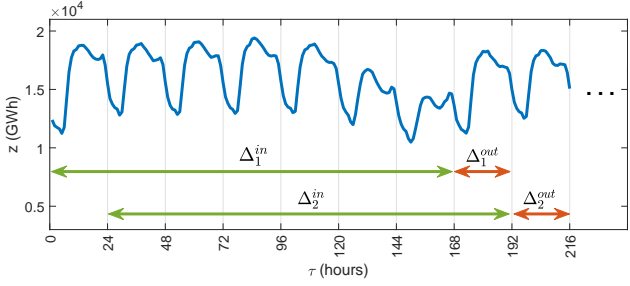}
	\caption{Moving windows used for preprocessing TS $z_\tau$.} 
	\label{figW}
\end{figure}

The windows are shifted by 24 hours to obtain subsequent input and output patterns (see Fig. \ref{figW}), which are defined as:

\begin{equation}
\begin{aligned}
\textbf{x}_1^{in} &= [x_1, ..., x_{168}], &\textbf{x}_1^{out} = [x_{169}, ..., x_{192}], \\
 \textbf{x}_2^{in} &= [x_{25}, ..., x_{192}], &\textbf{x}_2^{out} = [x_{193}, ..., x_{216}], \\
 ...
\end{aligned}
\label{eqpt}
\end{equation}
where the $t$-th pair of patterns represent deseasonalized, normalized and squashed TS sequences covered by the $t$-th pair of windows: 

\begin{equation}
x_\tau=\log{\frac{z_\tau}{\bar{z}_t \hat{s}_{t,\tau}} }
\label{eqxt}
\end{equation}
where $\tau \in \Delta^{in}_t \cup \Delta^{out}_t$, i.e. $\tau \in [24(t-1)+1, 24(t-1)+192]$,  $\bar{z}_t$ is the average TS value in the $t$-th input window, i.e. $\bar{z}_t=1/168 \sum_{24(t-1)+1}^{24(t-1)+168}z_\tau$, and $\hat{s}_{t,\tau}$ is the seasonal component determined using \eqref{eqls1} for recursive step $t$ (in this step $t$-th pair of patterns \eqref{eqpt} are used for RNN training).

TS sequences are squashed using a $\log$ function to prevent outliers from upsetting the learning process. Note that in \eqref{eqxt}, seasonal component $s_{t,\tau}$ is adapted for each $t$-th patterns in each training epoch. Thus the training set has a dynamic character. It is updated on-the-fly during learning. This process can be seen as the search for the optimal representation for RNN. 

To introduce more input information related to the forecasted sequence, the input patterns are extended as follows: 

\begin{equation}
\textbf{x}_t^{in'}= [\textbf{x}_t^{in},\, \hat{\textbf{s}}_t,\, \log_{10}(\bar{z}_t),\, \textbf{d}_t^{w},\, \textbf{d}_t^{m},\, \textbf{d}_t^{y}] 
\label{eqxp}
\end{equation} 
where 
$\hat{\textbf{s}}_t$ is a vector of 24 seasonal components predicted by ES for the output period $t$ reduced by 1, i.e. $\hat{\textbf{s}}_t = [\hat{s}_{t,\tau}-1]_{\tau=24(t-1)+169}^{24(t-1)+192}$,
$\textbf{d}_t^{w} \in \{0, 1\}^7, \textbf{d}_t^{m} \in \{0, 1\}^{31}$ and $\textbf{d}_t^{y} \in \{0, 1\}^{52}$ are binary one-hot vectors encoding day of the week, day of the month and week of the year for the forecasted day, respectively.

Vectors $\textbf{d}_t^{w}$ and $\textbf{d}_t^{y}$ inform about the location of the forecasted sequence in the weekly and yearly cycles, $\textbf{d}_t^{m}$ helps to deal with fixed-date public holidays, $\log_{10}(\bar{z}_t)$ informs about the level of the TS (squashing function matches the level range with the range of the other components of $\textbf{x}_t^{in'}$), and $\hat{\textbf{s}}_t$ introduces additional information about the daily variability.   

RNN is trained on training samples $(\textbf{x}_t^{in'}, \textbf{x}_t^{out})$ (updated in each recursive step $t$), and produces forecasts of the output patterns $\hat{\textbf{x}}_t^{out} = [\hat{x}_\tau]_{\tau \in \Delta^{out}_t}$ and their quantiles defining PIs. The postprocessing component converts these forecasts to real value forecasts using transformed equation \eqref{eqxt}:

\begin{equation}
\hat{z}_\tau=\exp (\hat{x}_\tau) \bar{z}_t\hat{s}_{t,\tau}
\label{eqzt}
\end{equation} 

\subsection{RNN Component}

\begin{figure}
	\centering
	\includegraphics[width=0.33\textwidth]{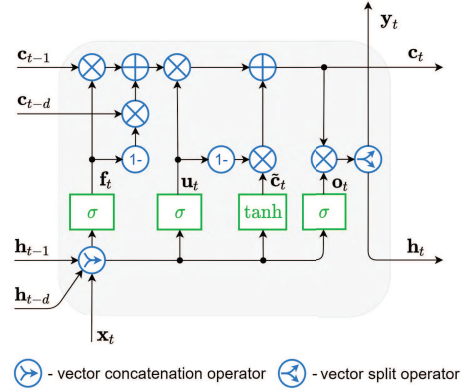}
	\caption{Proposed dilated RNN cell.} 
	\label{figS2}
\end{figure}

RNN employs a new type of gated recurrent cell, dilated RNN cell (dRNNCell), which is shown in Fig. \ref{figS2}. 
It is derived from the LSTM \cite{Hoch97} and GRU \cite{Cho14} cells. It is designed to operate as part of a multilayer dilated RNN \cite{Cha17} and as in \cite{Ben17} its output is split into "real output" $\textbf{y}_t$, which goes to the next layer, and a controlling output $\textbf{h}_t$, which is an input to the gating mechanism in the following time steps.

The cell uses two states, $c$-state (also called a cell state), which is close to the standard LSTM or GRU state, and $h$-state, which is the controlling state (also called a hidden state). At each time step $t$, the whole dRNNCell input is a concatenation of $\textbf{x}_t$, $\textbf{h}_{t-1}$ and $\textbf{h}_{t-d}$,  where $\textbf{x}_t$ is a standard input at time $t$ (either from a previous layer or an input to the RNN), $\textbf{h}_{t-1}$  is the most recent $h$-state, and $\textbf{h}_{t-d}$ is the delayed state ($d>1$). Both $c$- and $h$-states are saved in a list, to be used as delayed states. The size of the $c$-cell is equal to the summed sizes of $h$-state and $y$-output, i.e. $s_c=s_h+s_y$.

The dRNNCell uses the following gates: fusion ($f$), update ($u$), and output ($o$) gates. All the gates transform nonlinearly input vectors $\textbf{x}_t$, $\textbf{h}_{t-1}$ and $\textbf{h}_{t-d}$ using sigmoid function ($\sigma$). A candidate $c$-state, $\tilde{\textbf{c}}_t$, is produced by transforming input vectors using $\tanh$ nonlinearity. All nonlinear transformations of the input vectors are as follows: 

\begin{equation}
\textbf{f}_t= \sigma(\textbf{W}_f \textbf{x}_t+\textbf{V}_f\textbf{h}_{t-1}+\textbf{U}_f\textbf{h}_{t-d}+ \textbf{b}_f)
\label{eqp}
\end{equation} 
\begin{equation}
\textbf{u}_t= \sigma(\textbf{W}_u \textbf{x}_t+\textbf{V}_u\textbf{h}_{t-1}+\textbf{U}_u\textbf{h}_{t-d}+ \textbf{b}_u)
\label{eqf}
\end{equation} 
\begin{equation}
\textbf{o}_t= \sigma(\textbf{W}_o \textbf{x}_t+\textbf{V}_o\textbf{h}_{t-1}+\textbf{U}_o\textbf{h}_{t-d}+ \textbf{b}_o)
\label{eqo}
\end{equation} 
\begin{equation}
\tilde{\textbf{c}}_t= \tanh(\textbf{W}_c \textbf{x}_t+\textbf{V}_c\textbf{h}_{t-1}+\textbf{U}_c\textbf{h}_{t-d}+ \textbf{b}_c)
\label{eqg}
\end{equation} 
where $\textbf{W}$, $\textbf{V}$, $\textbf{U}$ are weight matrices, and $\textbf{b}$ are bias vectors.

The $c$-state is a weighted combination of past $c$-states and new candidate state $\tilde{\textbf{c}}_t$ computed in the current step:

\begin{equation}
\textbf{c}_t = \textbf{u}_t  \otimes  
\left(\textbf{f}_t \otimes \textbf{c}_{t-1}  +
\left(1-\textbf{f}_t\right) \otimes \textbf{c}_{t-d}\right) +
(1-\textbf{u}_t) \otimes \tilde{\textbf{c}}_t
\label{eqc}
\end{equation}
where $\otimes$ denotes the Hadamard product (element-wise product).

Update vector $\textbf{u}_t$ decides in what proportion the old and new information are mixed in the $c$-state, while fusion vector $\textbf{f}_t$ decides about the contribution of  recent and delayed $c$-states in the new state.

The controlling state and the output of the cell is calculated based on the new $c$-state and output gate as follows:
 
\begin{equation}
\textbf{h}'_t = \textbf{o}_t   \otimes  \textbf{c}_t 
\label{eqh1}
\end{equation} 
\begin{equation}
\textbf{y}_t= [h'_1, ..., h'_{s_y}]
\label{eqy}
\end{equation}
\begin{equation}
\textbf{h}_t= [h'_{s_y+1}, ..., h'_{s_y+s_h}]
\label{eqh}
\end{equation}

\begin{figure}
	\centering
	\includegraphics[width=0.48\textwidth]{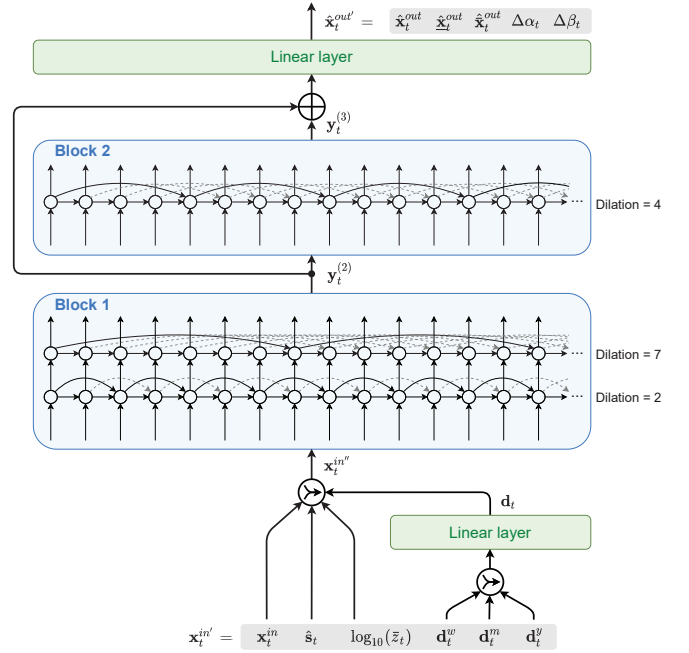}
	\caption{Proposed RNN architecture. The circles represent dRNNCells.} 
	\label{figRNN}
\end{figure}

The dRNNCell is a part of a multi-layer dilated RNN, which is composed of a number of blocks, each composed of one or more cells. In Fig. \ref{figRNN} there are two blocks, the first with two layers dilated 2 and 7, respectively, and the second, just with a single layer dilated 4. Dilated RNN architecture was introduced in \cite{Cha17} to tackle the three major challenges presented by RNN when learning on long sequences, i.e. complex dependencies, vanishing and exploding gradients, and efficient parallelization. 
Note that our new dRNNCell is fed by both recent ($t-1$) and delayed ($t-d$) states. Thanks to this, the cell uses directly information from both the previous step and a step distant in time. This can be useful for seasonal TS, where the relationships between the series elements have a cyclical character. These relationships can be modeled more accurately using dilated connections related to seasonality. To enable RNN to learn the temporal dependencies of different scales, we use multiple dilated recurrent layers stacked with hierarchical dilations. The proposed RNN uses ResNet-style shortcuts between blocks \cite{He16} to improve the learning process by preventing vanishing or exploding gradients.    

As can be seen from Fig. \ref{figRNN}, binary vectors encoding calendar data, $\textbf{d}_t^{w}$, $\textbf{d}_t^{m}$ and $\textbf{d}_t^{y}$, are embedded using a linear layer into $d$-dimensional continuous vectors $\textbf{d}_t$. This reduces input dimensionality and meaningfully represents sparse binary vectors in the embedding space. The embedding is learned along with the model itself.

The output layer in Fig. \ref{figRNN} is a linear one. It produces vector $\hat{\textbf{x}}_t^{out'}$ which is a concatenation of the forecasted output pattern, $\hat{\textbf{x}}_t^{out} = [\hat{x}_\tau]_{\tau \in \Delta^{out}_t}$, lower bounds of PI, $\hat{\underline{\textbf{x}}}_t^{out} = [\hat{\underline{x}}_\tau]_{\tau \in \Delta^{out}_t}$, upper bounds of PI, $\hat{\bar{\textbf{x}}}_t^{out} = [\hat{\bar{x}}_\tau]_{\tau \in \Delta^{out}_t}$, and corrections for smoothing coefficients, $\Delta{\alpha_t}$ and $\Delta{\beta_t}$:

\begin{equation}
\hat{\textbf{x}}_t^{out'}= [\hat{\textbf{x}}_t^{out},\, 
\hat{\underline{\textbf{x}}}_t^{out},\,
\hat{\bar{\textbf{x}}}_t^{out},\,
\Delta{\alpha_t},\,
\Delta{\beta_t}]
\label{eqout}
\end{equation}

\subsection{Loss Function}

To define the loss function we employ a pinball loss:

\begin{equation}
\rho(z, \hat{z}_q) =
\begin{cases}
(z-\hat{z}_q)q       & \text{if } z \geq \hat{z}_q\\
(z-\hat{z}_q)(q-1)  &\text{if } z < \hat{z}_q 
\end{cases}
\label{eqrho}
\end{equation}
where $z$ is an actual value, $\hat{z}_q$ is a forecasted value of $q$-th quantile, and $q \in (0, 1)$ is a quantile order.

The pinball loss is commonly used in quantile regression and probabilistic forecasting \cite{Tak06}. It helps us to determine the point forecasts and PIs, whose lower and upper bounds are expressed by quantiles of orders $\underline{q}$ and $\overline{q}$, respectively (e.g. $\underline{q}=0.05$ and $\overline{q}=0.95$).   

Our loss function has the following three components:
 
\begin{equation}
L_\tau =
\rho(z'_\tau, \hat{z}'_{q^*,\tau}) + \gamma(
\rho (z'_\tau, \hat{{z}}'_{\underline{q},\tau}) + 
\rho (z'_\tau, \hat{{z}}'_{\overline{q},\tau}))
\label{eqlss}
\end{equation}
where $q^*=0.5$ corresponds to the median, $z'_\tau = z_\tau / \bar{z}_t$ is a normalized actual TS value from the output window $\Delta^{out}_t$, 
$\hat{z}'_{q^*,\tau} = \exp(\hat{x}_\tau)\hat{s}_{t,\tau}$ is a forecasted value of $z'_\tau$, $\underline{q}, \overline{q}$ are the quantile orders for the lower and upper bounds of PI, respectively, 
$\hat{{z}}'_{\underline{q},\tau} = \exp(\hat{\underline{x}}_\tau)\hat{s}_{t,\tau}$  is a forecasted value of $\underline{q}$-quantile of $z'_\tau$, 
$\hat{{z}}'_{\overline{q},\tau} = \exp(\hat{\bar{x}}_\tau)\hat{s}_{t,\tau}$ is a forecasted value of $\overline{q}$-quantile of $z'_\tau$, and $\gamma \geq 0$ is a parameter controlling the impact of the components related to PI on the loss function, typically between 0.1 and 0.5. 

Note that loss function \eqref{eqlss} operates on the normalized TS values $z'_\tau$. This is because different TS can have different levels and normalization allows us to bring their errors expressed by \eqref{eqrho} to the same level, which is crucial in cross-learning. The forecasts of $z'_\tau$ in \eqref{eqlss} are calculated from \eqref{eqzt} excluding $\bar{z}_t$ to obtain normalized forecasts. These forecasts are based on the RNN outputs,  $\hat{\textbf{x}}_t^{out}, \hat{\underline{\textbf{x}}}_t^{out}$ and $\hat{\bar{\textbf{x}}}_t^{out}$, and ES outputs  $\{\hat{s}_{t,\tau}\}_{\tau \in \Delta^{out}_t}$.

The first component in \eqref{eqlss}, $\rho(z'_\tau, \hat{z}'_{q^*,\tau})$, represents a symmetrical loss for the forecasted value (normalized) while the second and third components, $\rho (z'_\tau, \hat{{z}}'_{\underline{q},\tau})$ and $\rho (z'_\tau, \hat{{z}}'_{\overline{q},\tau})$, represent asymmetrical losses for the quantiles. The asymmetry level, which determines PI, results from the quantile orders. Hyperparameter $\gamma$ determines the share of the three components in the loss function. 
For $\gamma = 1$, all the components have the same impact on the loss function. To increase the importance of the first component over the other two, we decrease the $\gamma$ value. Note that due to the three component parametrized loss function, we have the ability to optimize both the point forecasts and their PI. Moreover, the pinball loss gives us the opportunity to reduce the forecast bias by penalizing positive and negative deviations differently. When the model tends to have a positive or negative bias, we can reduce the bias by introducing $q^*$ smaller or larger than $0.5$, respectively (see \cite{Smy20, Dud21}).

            
    


\subsection{Mechanisms and Solutions for Performance Improvement}

The proposed model has the following mechanisms and solutions for improved performance: 

\begin{itemize}
\item dRNNCell with expanded states (recent and dilated ones). dRNNCell is able to model both short-term and long-term dependencies in TS. This feature is useful especially for STLF where TS express multiple seasonality. Temporal dependencies in this case have a cyclical character and can be modeled hierarchically using different dilations in different RNN layers.

\item Hybrid architecture combining ES and RNN. ES extracts dynamically the main components of each individual TS and enables appropriate TS representation for RNN. A multiple dilated stacked RNN architecture is able to deal with complex TS expressing multiple seasonality. The two components, ES and RNN, are optimized simultaneously by the same optimization algorithm. This fine-tunes RNN weights as well as ES smoothing coefficients. So the resulting forecasting model, including dynamic data preprocessing, is optimized as a whole. 

\item Cross-learning. The model is global. Learning across many TS enables it to capture the  shared features and components of the TS. Cross-learning is a type of multi-task learning \cite{Car97} which is known to be an effective method of improving generalization by using the domain information contained in the training samples of related tasks as an inductive bias. Moreover, cross-learning greatly speeds up the learning of deep architectures. 

\item A dynamic training set for RNN. The training samples are updated on-the-fly during learning. The optimal representation of TS is searched for to ensure the best predictive performance of the model.

\item To delay the onset of the over-training, the starting point of training is sampled, so the same TS is likely to look slightly differently each time a batch is formed.  

\item Three component, parametrized pinball loss function. This enables the model to optimize both the point forecasts and their PIs. Moreover, it enables the forecast bias to be reduced.   
\item Ensembling, which is a powerful regularization technique. This exploits the beneficial effects of combining forecasts \cite{Chan18}, improves accuracy and stability compared to a single learner, and enhances the robustness, thus mitigating model and parameter uncertainty \cite{Pet18}.

\end{itemize}

\section{Experimental Study}

In this section, we apply the proposed ES-dRNN model to STLF and compare its performance with that of other models including statistical and ML ones. We test the models on real-world data comprising hourly electricity demand TS for 35 European countries from the period 2016-2018 (source - ENTSO-E repository www.entsoe.eu/data/power-stats/). The TS were described and analysed in Section II. They differ substantially in levels, trends, dispersion and daily shapes. Thus, the data provides a variety of TS with different properties, which translates into a more reliable test for the forecasting models.

A one day-ahead forecasting problem is considered. We optimize ES-dRNN using the data from 2016 and 2017 (data for 2016 for Albania is unavailable). The model forecasts the daily load profile for each day of 2018 for each of the 35 countries with the exception of Estonia and Italy for which data for the last month of 2018 is unavailable, and Latvia for which data for the last two months of 2018 is unavailable.


\subsection{Optimization and Training Procedures}

During each epoch a number of updates is executed, guided by the average error accumulated by executing $l_o$ (e.g. 50) forward steps, moving by one day, on a batch. The starting point is chosen randomly; the batches include random $b$ series. The model is trained using Adam optimizer.

The $d$-dilated dRNNCell operates as described above only after $d$ steps, because only after $d$ steps are the delayed states available. Additionally, the Holt-Winters formulas require at least twice the seasonality steps to stabilize, so the system uses several weeks ($w_o$) at the beginning of each batch as a warm-up period, during which all the ES and RNN calculations take place, with the exception of the training errors, which are not calculated.
Similarly, an even longer warm-up period $w_s$ is applied when producing the test results.

An epoch is usually defined as using all the training data once. Our definition here is based on the number of updates or processed batches, as during training we step $l_o$ times on a batch (with random assignment of series) and for each batch execute a single update based on the average error. Our aim is to define the epoch as the number of updates which brings in a meaningful change in the learning process, and because the data set contains a small number of series, a single epoch is actually composed with $n_o$ number "sub-epochs", defined in the traditional fashion as one scan of all available data. An additional factor is the batch size: when it grows, the number of updates per sub-epoch diminishes, so the number of the sub-epochs needs to grow. However, in our experience the linear growth is too fast, and risks overfitting within a single epoch, so finally we use the following formula 
\begin{equation}
n_o=\min{\left(1,\left(\frac{Nb}{L}\right)^p\right)}
\end{equation}
where $N$ is the maximum number of updates per epoch, $b$ is the current batch size, $L$ is the number of TS in the data set, and $p$ is a hyperparameter, between 0 and 1, which by experimentation is set to 0.7.

The model hyperparameters were  selected as follows:
\begin{itemize}
\item Number of epochs. In early testing we established that 9 epochs is usually sufficient to reach a plateau of accuracy.

\item Number of TS in the batch. We use the schedule of increasing batch sizes and decreasing learning rates proposed in \cite{Smi18}. We start with a small batch size of 2, and increase it, although only once, due to the small number of series, to 5 at epoch 4. 
\item Learning rates. Decreasing learning rates has a similar, if not the same, effect as increasing the batch size: it allows the validation error to be  further reduced. We use the following schedule: $3\cdot10^{-3}$ (epochs 1-4), $10^{-3}$ (epoch 5), $3\cdot10^{-4}$ (epoch 6), $10^{-4}$ (epochs 7-9), 

\item Size of the $c$-state, $h$-state and $y$-output: $s_c=100$, $s_h=40$, $s_y=60$.
Increasing the size of cells causes a quadratic increase in the number of parameters. Larger models may be beneficial for larger data sets. The values above were obtained by experimentation starting with $s_c=50$, $s_h=20$ and doubling it 3 times.

\item The RNN architecture and dilations. We use three layers: two dilated by 2 and 7 in the first block, and a single layer dilated by 4 in the second block (see Fig. \ref{figRNN}). We arrived at this layout partly by heuristics and partly by experimentation. We started with two layers, the first with a minimum dilation 2 (because the previous values, dilated by 1, are always used by the cells), and the second layer dilated by 7, to match them with the main weekly seasonality. Then we tried to add a third layer, and to avoid the vanishing gradient problem it had to be in a new block. We chose dilation 4, because together with 2 and 7, this makes an almost perfect geometric series, as advocated by \cite{Cha17}. But of course we also experimented with larger dilation 11 and 14 in the third layer, but the results were worse, 
which is as expected, because our horizon is just 1 day. Among the alternatives arrangements for the three layers, three blocks of a single layer each dilated 2, 4 and 7 works equally well.

\item Loss function parameters. As described in Section III D, the pinball loss function was utilized, with three different quantile values $q$,  to achieve quantile regression for 0.5, 0.05, and 0.95. The actual values for $q^*$, $\underline{q}$, and $\overline{q}$ were slightly different: 0.49, 0.035, 0.96. These values were arrived at by experimentation, reducing the bias of the center value, and fine-tuning the percentage of exceedance for PIs. The $\gamma$ parameter was 0.3, set, without experimentation, following a rule of thumb that states that the sum of PIs losses should be around 10-20\% of the center value loss, reflecting the usual higher importance applied to the center loss. However, it is not a sensitive parameter, and leaving it at 1 would not make much difference.

\item Initial smoothing coefficients: $I\alpha=-3.5$, $I\beta=0.3$. These were arrived at by observing, during early runs of the training,  the direction and size of average adjustments to both smoothing coefficients, by the NN. As expected, the level is quite stable, so the smoothing coefficient tends to be close to zero, and therefore the $I\alpha$ is a relatively large negative number. Seasonality, on the other hand, is likely to change more, and this is confirmed by larger, typically above 0.5, smoothing coefficients, starting from $I\beta=0.3$.

\item Lengths of TS sequences in the optimization mode: $l_o=50$. The longer the sequence, the more smooth the gradient should be, but at the same time the system may "see" the same parts of a series too often and can overtrain quickly. Additionally, experimentation suggested that smaller values of 20 and 30 brought worse results.

\item Lengths of training $w_o$ and testing $w_s$ warm-up periods: 3 and 13 weeks, respectively. The training warm-up period $w_o$ needs to be just slightly longer than 2 weeks, twice the seasonality size, for ES to stabilize. The number of warming-up steps should also be larger than the smallest dilation, but this is just 2 steps (days), so this second condition is not important here. The length of the testing warm-up period was chosen mostly due to prior experience that stepping through 2-3 months of typical business TS is enough for the system to fully "zero-in" on a particular series.

\item Embedding size of the calendar variables: 4. This value was arrived at firstly by the expectation that the one-hot encoded input of size 7+31+52 should be able to be converted to an order of magnitude smaller floating point vector, and then by experimentation.

\item Ensemble size: $E=100$, although a size as small as 5 is often sufficient.

\end{itemize}

\subsection{Baseline Models}

We compare our ES-dRNN in terms of accuracy with the baseline models outlined below:

\begin{itemize}
\item Naive -- naive model in the form: the forecasted demand profile for day $i$ is the same as the profile for day $i-7$
\item ARIMA -- autoregressive integrated moving average model \cite{Dud15},
\item ES -- exponential smoothing model \cite{Dud15},
\item Prophet -- modular additive regression model with nonlinear trend and seasonal components \cite{Tay18},
\item k-NNw -- weighted $k$-nearest neighbour method \cite{Dud15},
\item FNM -- fuzzy neighborhood model \cite{Dud15}
\item N-WE -- Nadaraya–Watson estimator \cite{Dud15}
\item GRNN -- general regression NN \cite{Dud16a},
\item MLP -- perceptron with a single hidden layer and sigmoid nonlinearities \cite{Dud16a},
\item SVM -- linear epsilon insensitive support vector machine ($\epsilon$-SVM) \cite{Pel21},
\item LSTM -- long short-term memory \cite{Pel20},
\item ANFIS -- adaptive neuro-fuzzy inference system \cite{Pel18},
\item MTGNN -- graph NN for multivariate TS forecasting \cite{Wu20}.
\end{itemize}

The baseline models include classical statistical models (ARIMA, ES), new statistical models (Prophet), nonparametric pattern-based ML models (k-NNw, FNM, N-WE), classical ML models (MLP, GRNN, SVM, ANFIS) and new recurrent and deep NN architectures (LSTM, MTGNN).

\subsection{Results}

In this section, we report the results for our proposed model in two variants: as an individual model, denoted by ES-dRNN, and as an ensemble of $E$ ES-dRNNs, denoted by ES-dRNNe.

Table \ref{tabEr} shows the results of forecasting averaged over all 35 countries, i.e. mean absolute percentage error (MAPE), median of APE (MdAPE), interquartile range of APE (IqrAPE), root mean square error (RMSE), mean PE (MPE), and standard deviation of PE (StdPE). MdAPE measures the average error without the influence of outliers, while RMSE is especially sensitive to outliers as a square error. MPE measures the forecast bias. Note the lowest values for MAPE, MdAPE and RMSE for ES-dRNNe and the second lowest for ES-dRNN. Our models also produce the least dispersed predictions compared to the baseline models (IqrAPE $\leq 2.25$). 



\begin{table}[]
	\caption{Forecast results.}
	\begin{tabular}{lcccrrc}
		\toprule
		& MAPE  & MdAPE & IqrAPE & RMSE  & MPE   & StdPE \\
		\midrule    
    Naive & 5.08  & 4.84  & 3.32  & 704.34 & -0.26 & 7.91 \\
    ARIMA & 3.30  & 3.01  & 3.00  & 475.09 & -0.01 & 5.31 \\
    ES   & 3.11  & 2.88  & 2.73  & 439.26 & 0.01  & 5.13 \\
    Prophet & 4.53  & 4.32  & 3.03  & 619.39 & -0.13 & 6.82 \\
    k-NNw & 2.50  & 2.31  & 2.30  & 335.13 & -0.11 & 4.26 \\
    FNM   & 2.50  & 2.30  & 2.29  & 334.08 & -0.11 & 4.27 \\
    N-WE  & 2.49  & 2.28  & 2.30  & 332.49 & -0.13 & 4.26 \\
    GRNN  & 2.48  & 2.28  & 2.27  & 332.91 & -0.11 & 4.25 \\
    MLP   & 3.05  & 2.78  & 2.94  & 419.01 & -0.04 & 5.07 \\
    SVM   & 2.55  & 2.29  & 2.52  & 357.24 & -0.13 & 4.37 \\
    LSTM  & 2.76  & 2.57  & 2.52  & 381.76 & \textbf{0.02} & 4.47 \\
    ANFIS   & 3.65  & 3.17  & 3.66  & 507.08 & -0.10 & 6.43 \\
    MTGNN & 2.99  & 2.74  & 2.69  & 405.18 & -0.47 & 4.85 \\
    ES-dRNN & 2.36  & 2.15  & 2.25  & 326.06 & -0.20 & 3.89 \\
    ES-dRNNe & \textbf{2.23} & \textbf{2.02} & \textbf{2.15} & \textbf{306.94} & -0.20 & \textbf{3.75} \\
		\bottomrule
	\end{tabular}
	\label{tabEr}
\end{table}

\begin{figure}
	\centering
	\includegraphics[width=0.28\textwidth]{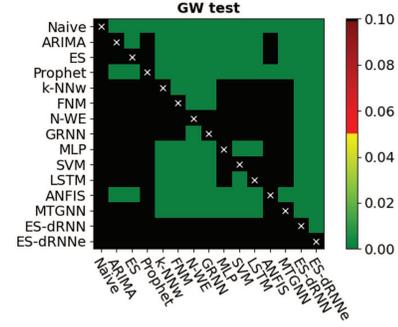}
	\caption{Results of the Giacomini-White tests.} 
	\label{figGW}
\end{figure}

The winning performance of ES-dRNNe and ES-dRNN was confirmed using a pairwise one-sided Giacomini-White test (GM test) for conditional predictive ability \cite{Gia06}. We used an implementation of the GW test in the multivariate variant from https://github.com/jeslago/epftoolbox \cite{Lag21}. Fig. \ref{figGW} shows results of the GW test, i.e. a heat map representing the obtained $p$-values. The closer they are to zero
the significantly more accurate the forecasts produced by the model on the $X$-axis are than the forecasts produced by the model on the $Y$-axis. The black color is for $p$-values larger than 0.10 indicating rejection of the hypothesis that the model on the $X$-axis is more accurate than the model on the $Y$-axis.
Note that both ES-dRNNe and ES-dRNN performed significantly better in terms of accuracy than all the other comparative models.  

\begin{figure}
	\centering
	\includegraphics[width=0.48\textwidth]{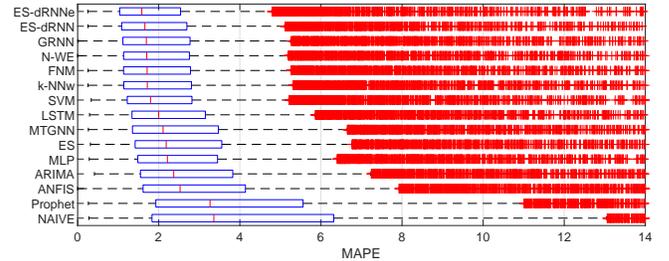}
	\caption{Boxplots for daily MAPE.} 
	\label{figBx}
\end{figure}

\begin{figure}
	\centering
	\includegraphics[width=0.48\textwidth]{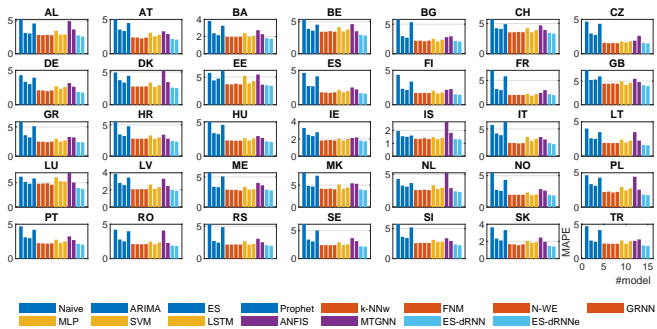}
	\caption{MAPE of the models for each country.} 
	\label{figPan}
\end{figure}

\begin{figure}
	\centering
	\includegraphics[width=0.48\textwidth]{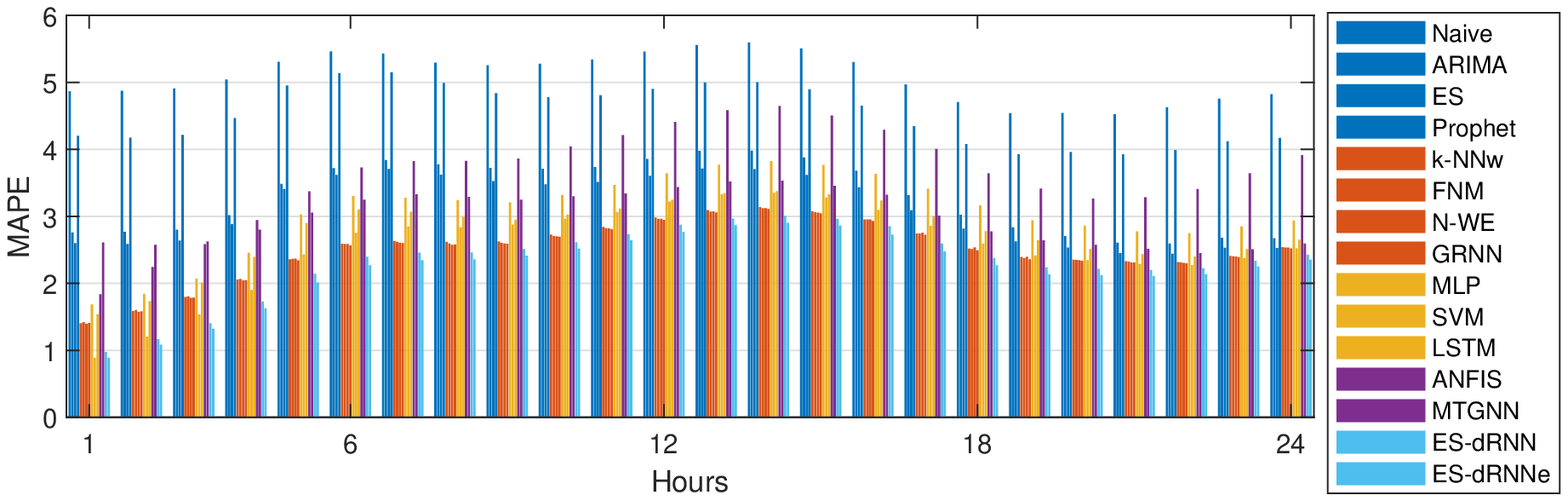}
	\includegraphics[width=0.20\textwidth]{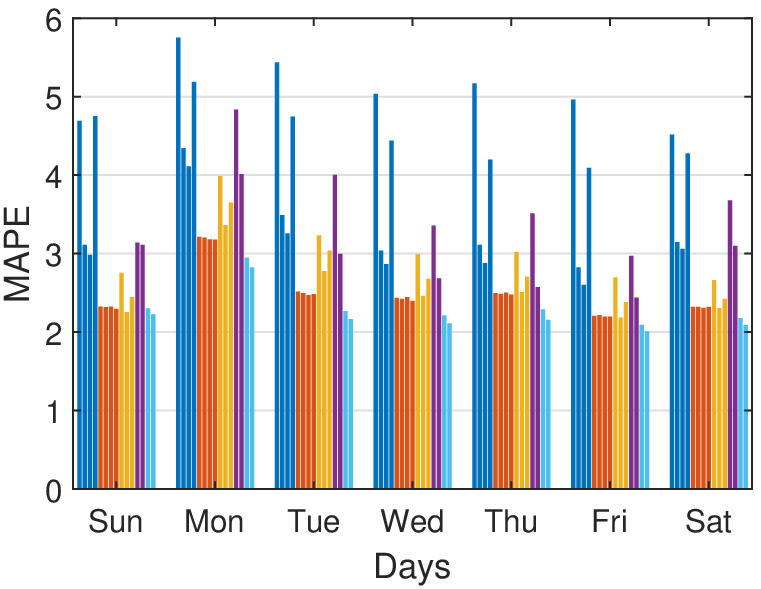}
	\includegraphics[width=0.28\textwidth]{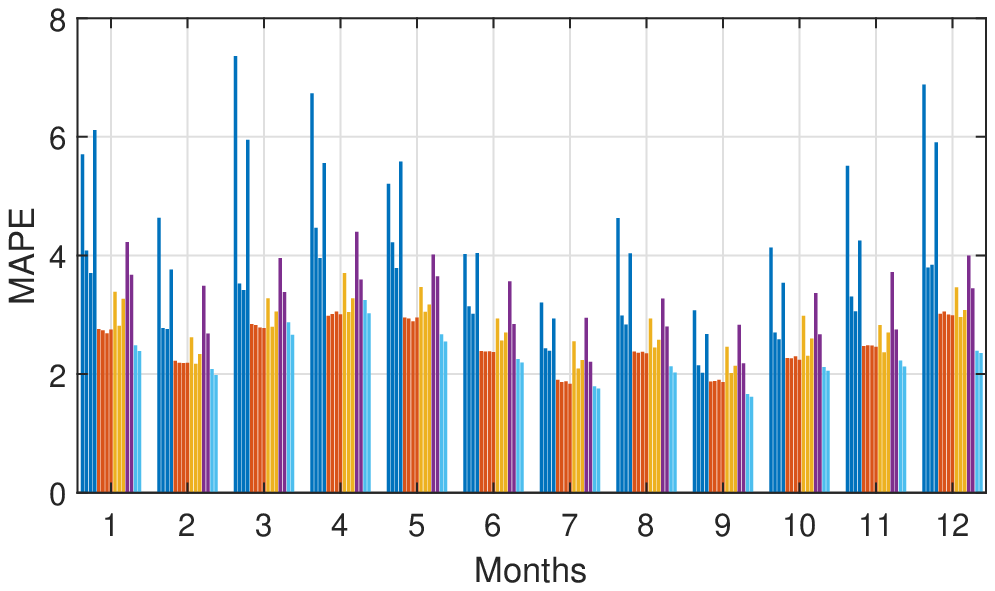}
	\caption{MAPE for each hour of the day, each day of the week and each month.} 
	\label{figHDM}
\end{figure}

More detailed results are shown in Figs. \ref{figBx}-\ref{figHDM}. From Fig. \ref{figBx} we can assess distribution of the daily MAPE. Note the smallest medians and the most compact distributions for our model, which is ahead of the group of nonparametric ML models designed specifically for STLF. Fig. \ref{figPan} shows MAPE for individual countries. Our model was the most accurate for all countries except for France, where it was beaten by SVM, and Montenegro, where it was beaten by GRNN. Fig. \ref{figHDM} demonstrates the average errors for each hour of the day, each day of the week and each month of the test period (2018). Note that in each case errors for ES-dRNNe and ES-dRNN are among the lowest.           

\begin{figure}
	\centering
	\includegraphics[width=0.158\textwidth]{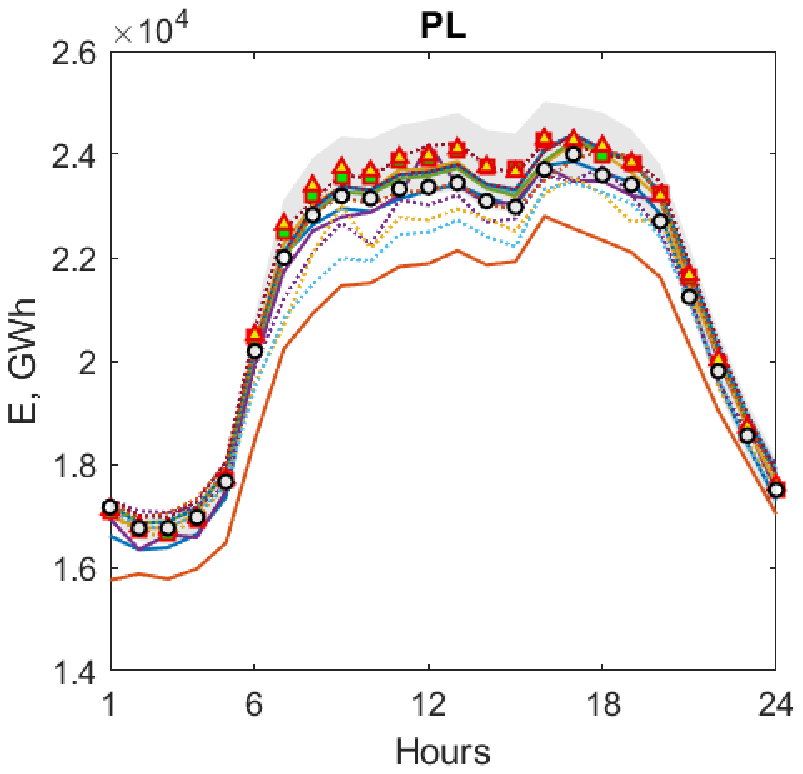}
	\includegraphics[width=0.158\textwidth]{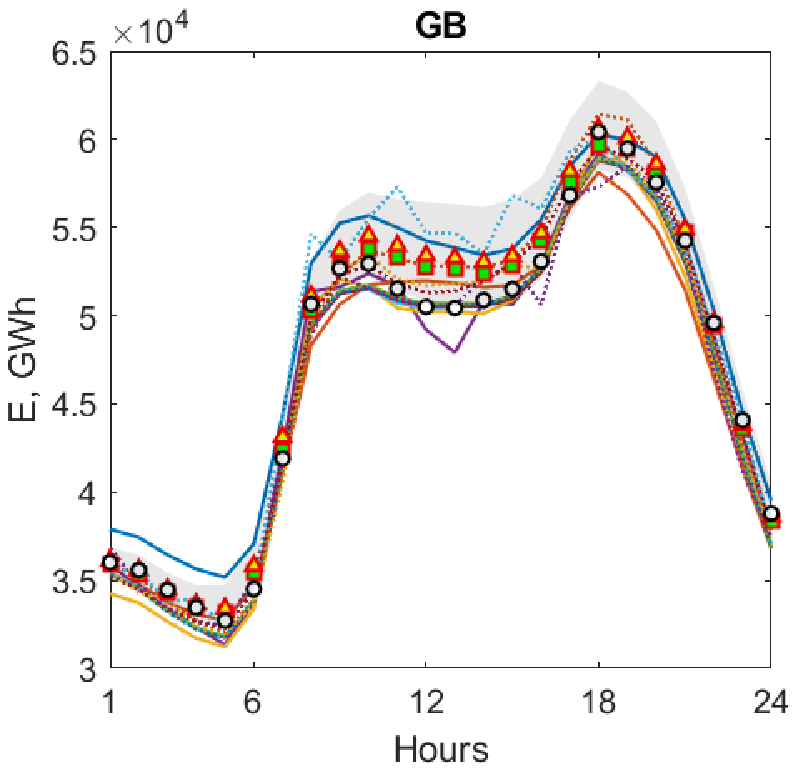}
	\includegraphics[width=0.158\textwidth]{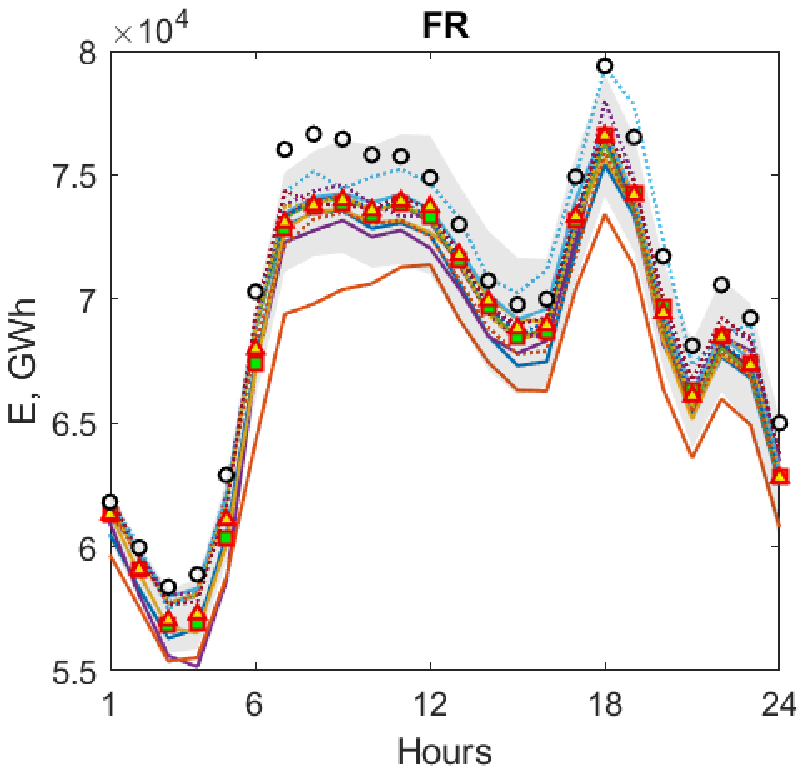}
	\includegraphics[width=0.158\textwidth]{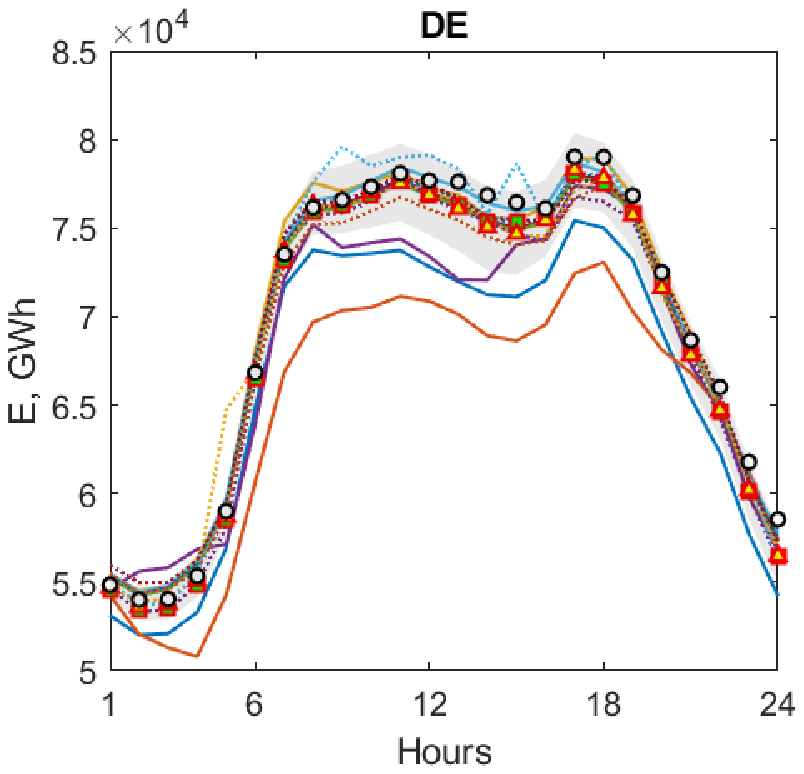}
	\includegraphics[width=0.158\textwidth]{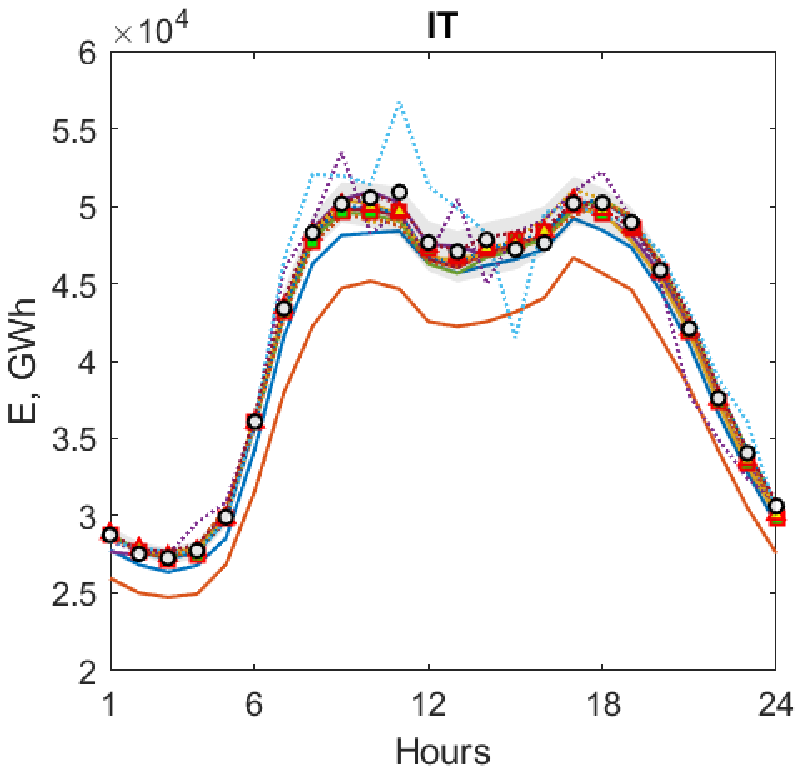}
	\includegraphics[width=0.158\textwidth]{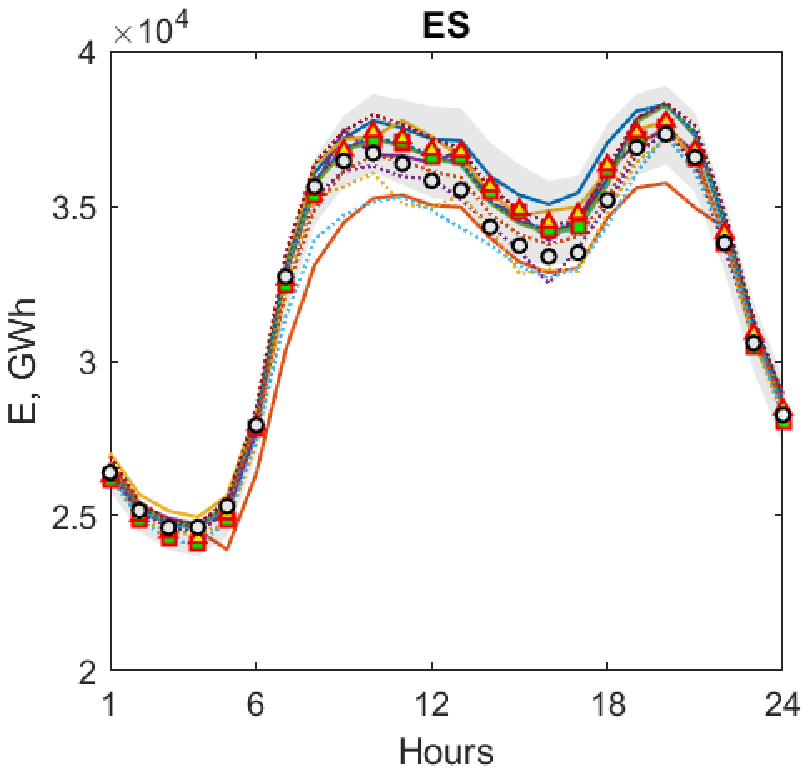}
	\includegraphics[width=0.48\textwidth]{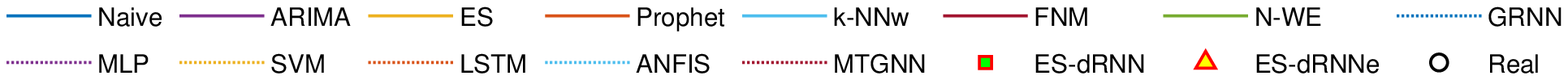}
	\caption{Examples of the forecasted daily profiles. 90\% PIs for ES-dRNNe are shown as gray-shaded areas.} 
	\label{figPro}
\end{figure}

Fig. \ref{figPro} shows some examples of the daily profile forecasts produced by our models and baseline models. From this figure we can assess the fitting of the models to the real data. 
In Fig. \ref{figPro} the PIs predicted by ES-dRNNe are also shown. To assess the PIs we calculate for each country the number of forecasted values in PIs, below PIs and above PIs. We achieved: $90.14\% \pm 2.43\%$, $4.88\% \pm 1.29\%$ and $4.98\% \pm 1.41\%$, respectively. These values corresponds to our assumed 90\% PIs with lower and upper bounds $\underline{q}=0.05$ and $\overline{q}=0.95$, respectively.




\subsection{Ablation Study}

The proposed ES-dRNN has several components and mechanisms to increase its predictive power for STLF. In the ablation study we test the performance of the reduced model. We reduce the model as follows:

\begin{description}
\item[\textbf{Ab1}] ES component is removed. dRNN learns on the normalized, but not deseasonalized TS.
Seasonality vector $\hat{\textbf{s}}_t$ is excluded from input pattern \eqref{eqxp}.

\item[\textbf{Ab2}] ResNet-style shortcut between blocks 1 and 2 is removed. ES-dRNN learns without residual connection. The linear output layer is fed directly with the output vector of block 2, $\textbf{y}_t^{(3)}$ (see Fig. (\ref{figRNN})).  

\item[\textbf{Ab3}] dRNNCell without fusion gate is used. The delayed $c$-state is used, if available, otherwise recent, $t-1$, state is used.

\item[\textbf{Ab4}] dRNNCell without dilated states is used. It is fed with only $t-1$ states.

\item[\textbf{Ab5}] dRNNCell without recent states is used. It is fed with only $t-d$ states.

\item[\textbf{Ab6}] LSTM cell is used instead of dRNNCell. The cell is simplified, without delayed connections.

\item[\textbf{Ab7}] Level input $\log_{10}(\bar{z}_t)$ is excluded from input pattern \eqref{eqxp}.

\item[\textbf{Ab8}] Inputs $\log_{10}(\bar{z}_t)$, $\textbf{d}_t^{w}$, $\textbf{d}_t^{m}$ and $\textbf{d}_t^{y}$ are excluded from input pattern \eqref{eqxp}. No input information about the TS level and current location in the weekly, monthly and yearly cycles is introduced to dRNN.

\item[\textbf{Ab9}] Inputs $\log_{10}(\bar{z}_t)$, $\textbf{d}_t^{w}$, $\textbf{d}_t^{m}$, $\textbf{d}_t^{y}$ and $\hat{\textbf{s}}_t$ are excluded from input pattern \eqref{eqxp}. dRNN is fed with only input pattern $\textbf{x}_t^{in}$. Additional input information such as TS level, current seasonality and calendar data is removed.

\item[\textbf{Ab10}] Embedding is excluded. Extended input vector \eqref{eqxp} is directly introduced on block 1. Input linear layer transforming calendar one-hot vectors into continuous embedding vector is removed.  

\end{description}

As can be seen from Table \ref{tabAb}, the lowest errors were achieved by the full model. Any reduction in the model leads to an increase in the forecast error.  

\begin{table}[]
	\setlength{\tabcolsep}{1.5pt}
	\caption{Errors for full and reduced model.}
	\begin{tabular}{crrrrrrrrrrr}
		\toprule
		& Full  & Ab1   & Ab2   & Ab3   & Ab4   & Ab5   & Ab6   & Ab7   & Ab8   & Ab9   & Ab10 \\
		\midrule    
    MAPE  & \textbf{2.227} & 2.267 & 2.247 & 2.240 & 2.231 & 2.264 & 2.245 & 2.232 & 2.286 & 2.345 & 2.291 \\
    RMSE  & \textbf{306.9} & 310.2 & 310.5 & 309.1 & 307.0 & 310.9 & 307.2 & 307.9 & 313.1 & 321.8 & 315.4 \\
 		\bottomrule
	\end{tabular}
	\label{tabAb}
\end{table}

\subsection{Discussion}

The experimental study proves that both variants of ES-dRNN clearly outperform all other models in terms of accuracy. The wide range of STLF problems on which we have tested the algorithms increases our confidence in this conclusion.
The distinguishing feature of our model from other ML baseline models is that it produce both point forecasts and PIs with a specified probability coverage. Thus the user gains additional information about the uncertainty of the prediction.

The proposed model is equipped with several mechanisms and solutions for performance improvement (see Section III E). Many of them were tested and proved their effectiveness on forecasting problems from diverse domains (see the winning submission to the M4 competition \cite{Smy20} and the model for monthly electricity demand forecasting \cite{Dud21}). Other components and mechanisms were designed in this study especially for STLF to deal with complex seasonality and short and long-term dependencies. As the ablation study has shown, the most important of them turned out to be: extended input information including daily variability, TS level and the calendar variables (Ab9), embedding of the calendar data (Ab10), and ES component (Ab1). Removing these components and mechanisms worsen the results the most.

We confirmed the beneficial effect of ensembling on increasing the accuracy of the forecasting model. The errors for the ensemble version were lower than those for the individual version by 5.51\%  for MAPE, 6.05\% for MdAPE and 5.86\% for RMSE. In our approach, ensembling does not require additional effort related to the selection of additional hyperparameters, e.g. controlling the diversity of individual learners. The diversity of learners is provided by the random initialization of the model parameters. However, controlling diversity could be an additional way  of improving performance.

The proposed ES-dRNN is more complex than the baseline statistical and ML models. It has a larger number of parameters (around 229K) and  hyperparameters to tune. However, the development of the model did not require long processing times nor any special hardware - it was done on a desktop-class computer, without GPU. A single training takes less than one hour, and can be done in parallel using a number of workers, allowing the results of the ensemble to be calculated immediately. Our experience with similar models allowed us to limit the number of hyperparameters combinations and code modifications. It is also worth noting that, once this kind of model is trained, the NN weights can be saved, and a serving program that uses them can be built. Such a serving program can forecast automatically and rapidly (in a matter of seconds) all the TS, when fed with new data. Retraining does not need to happen often, perhaps just twice a year. 
We worked with a relatively small data set covering 2-3 years. Considering strong yearly seasonality, extensive tuning of the hyperparameters, e.g. in order to completely remove bias, would likely have led to overfitting, so we purposefully avoided it.

\section{Conclusion}

In this paper, we proposed and empirically validated a new hybrid hierarchical architecture for STLF -- ES-dRNN. 
The empirical study of STLF for 35 European countries showed that our ES-dRNN had a significantly better performance than statistical and ML methods. It clearly outperformed its competitors in terms of accuracy. Its success is due to its unique hybrid architecture which combines ES and RNN. To deal with multiple seasonalities and short and long-term dependencies in TS we designed a new dilated recurrent cell and multiple dilated stacked RNN architecture. ES extracts dynamically the main components of each individual TS and enables appropriate TS representation for RNN. Due to the simultaneously learning of both ES and dRNN components the model is optimized as a whole. Cross-learning, i.e., learning on multiple TS, enables ES-dRNN to capture the shared features and components of each individual TS.      

The model produces point forecasts and PIs to express the forecast uncertainty. To optimize both, we introduced a new three component, parametrized loss function. This  also allows the forecast bias to be controlled.
A major advantage of ES-dRNN is its ability to deal with raw TS without any kind of preprocessing  such as decomposition or stationarization. All necessary data processing takes place inside the model. 

The high expressive power of the proposed model to solve nonlinear stochastic forecasting problems with complex seasonalities and significant random fluctuations has encouraged us to apply it to solve other complicated forecasting problems. This will be the focus of our future work.  



%


\ifCLASSOPTIONcaptionsoff
  \newpage
\fi



%





\end{document}